%% file: arxiv.tex
\documentclass[11pt]{article}
\usepackage{sustyle}
\usepackage{graphicx}
\usepackage{enumitem}
\usepackage{booktabs}

\usepackage{amsmath}

\usepackage{hyperref}

\usepackage{orcidlink}

\newcommand{\xgen}{x^{(\text{gen})}}

\newcommand{\U}{v}

\newcommand{\Di}{D^{(i)}}

\newcommand{\logo}{FlickrLogo-27}

\newcommand{\SRS}{\texttt{SRS}}

\newcommand{\playerset}{\mathcal{P}}

\newcommand{\aicomp}{\texttt{AI-Dev}}
\newcommand{\ShareData}{ \beta_{\text{data}} }
\newcommand{\loo}{\mathrm{loo}}

\include{macro}

\begin{document}

\title{An Economic Solution to Copyright Challenges of Generative AI}




\author{Jiachen T. Wang\textsuperscript{1} \and Zhun Deng\textsuperscript{2} \and Hiroaki Chiba-Okabe\textsuperscript{4} \and Boaz Barak\textsuperscript{3}\footnote{Currently also at OpenAI. Work done while at Harvard University.}
\and \\[-.8em] Weijie J.~Su\textsuperscript{4}}

\date{}


\maketitle

{\centering
\vspace*{-0.3cm}
\noindent\textsuperscript{1} Princeton University\\
\textsuperscript{2} Columbia University\\
\textsuperscript{3} Harvard University\\
\textsuperscript{4} University of Pennsylvania\par\bigskip 
\date{September 3, 2024}\par
}

\begin{abstract}
Generative artificial intelligence (AI) systems are trained on large data corpora to generate new pieces of text, images, videos, and other media. There is growing concern that such systems may infringe on the copyright interests of training data contributors. To address the copyright challenges of generative AI, we propose a framework that compensates copyright owners proportionally to their contributions to the creation of AI-generated content. The metric for contributions is quantitatively determined by leveraging the probabilistic nature of modern generative AI models and using techniques from cooperative game theory in economics. This framework enables a platform where AI developers benefit from access to high-quality training data, thus improving model performance. Meanwhile, copyright owners receive fair compensation, driving the continued provision of relevant data for generative model training. Experiments demonstrate that our framework successfully identifies the most relevant data sources used in artwork generation, ensuring a fair and interpretable distribution of revenues among copyright owners.

\end{abstract}

\section{Introduction}

Recent advancements in generative artificial intelligence (AI) have profoundly impacted the creative industries, ushering in an era of AI-generated content in literature, visual arts, and music. Trained on vast datasets of human-generated material, generative AI models such as large language models and diffusion models can now produce content with a sophistication that rivals---and may potentially displace---the works of human artists \cite{rombach2022high, bubeck2023sparks, henderson2023foundation}. This burgeoning capability raises crucial questions about the legal and ethical boundaries of creative authorship, particularly concerning copyright infringement by generative models \cite{sag2023copyright,samuelson2023generative}. Consequently, several AI companies are currently involved in lawsuits over allegations of producing content that potentially infringes on copyrights \cite{samuelson2023generative, grynbaum2023times}.

Efforts to mitigate the tension between owners of copyright in the training data and AI developers have emerged, mostly involving modifications to generative model training or inference to reduce the likelihood of generating infringing outputs \cite{vyas2023provable, chu2023protect, shan2023glaze, he2024fantastic, chibaokabe2024tackling}. However, these modifications may compromise model performance due to either the exclusion of high-quality, copyrighted training data from training or restrictions on content generation \cite{lee2023talkin}. The complexity and ambiguity of copyright law add another layer of difficulty, blurring the line between infringing and non-infringing outputs. The resulting uncertainty could lead to a significant waste of resources on both sides while these issues are debated in courts \cite{samuelson2023generative}.

Rather than restricting AI developers' use of copyrighted data, we propose establishing a mutually beneficial revenue-sharing agreement between AI developers and copyright owners. This proposal echoes an argument recently advocated in economics \cite{brynjolfsson2023generative}. 
However, a major challenge in developing
a revenue-sharing model for generative AI, in contrast to conventional cases of sharing between digital platforms and independent content creators \cite{deng2023computational}, lies in the complexity of training generative models on diverse data sources. This results in the ``black-box'' nature of model training and content generation, making the traditional, straightforward pro rata methods unsuitable \cite{lei2023pro}.

In this paper, we introduce a simple framework that appropriately compensates copyright owners for using their copyrighted data in training generative AI systems based on the cooperative game theory, thereby directly addressing the intricacies of copyright challenges. 
Our framework does not require modifying the inference process and preserves the full capabilities of generative models. We propose the royalty distribution model for sharing revenue with copyright owners by leveraging the probabilistic nature of generative models: the log-likelihood of generating the user-chosen content is used to measure the utility of the training data. This utility measure captures the capabilities of the model in satisfying users' needs. 
Royalties are subsequently distributed among the copyright owners in accordance with their contributions, which are analytically determined using the theory of Shapley value \cite{shapley1953value}. By aligning compensation with these quantifiable contributions, our framework ensures interpretability in the distribution of royalties, thereby fostering innovation in AI while guaranteeing a fair share of benefits to all copyright holders.

\begin{figure}[t]
    \centering
    \includegraphics[width=\linewidth]{images/flow_7.pdf}
    \caption{
    Overview of our method. 
    (a) The artists provide their copyrighted artworks as (part of) the training data for the generative AI model. 
    (b) A user prompts the generative AI and obtains a new artwork. 
    (c) We assess the contribution of each artist to the AI-generated artwork using the Shapley Royalty Share, which determines their compensation.
    }
    \label{fig:flowchart}
\end{figure}

\section{The Shapley Royalty Share Framework}

Our framework takes two steps to tackle copyright issues associated with generative AI models. The first step is to evaluate the utility of the model trained on every possible subset of the entire dataset. Intuitively, the utility of the data subset would be large if this model could generate with a great chance the same AI-generated content (e.g., an artwork) as the deployed model, which is trained on the entire dataset. The second step is to determine any participating copyright owners' rightful share based on the utilities from the first step, using tools from cooperative game theory. 
Loosely speaking, a copyright owner's share would be large if the utility tends to increase by including its data in the model training.

\paragraph{Utilities of different data source combinations}
Let there be $n$ copyright owners and the $i$th owns the copyright of training data $D^{(i)}$, where $i \in N := \{1, \ldots, n\}$. The deployed model is trained on the entire dataset $D :=  D^{(1)} \cup \cdots \cup D^{(n)}$ and generates a content $\xgen$. Consider a counterfactual model that is trained on a subset of training data, $\cup_{i \in S} \Di$, where $S \subset N$ denotes a subset of data owners. 
The utility of the counterfactual model could be best reflected by its likelihood of generating the same content $\xgen$ as the deployed model. 
Let $p_S(\cdot)$ denote the probability density function of this counterfactual model. 
We define the utility of this model for content $\xgen$ as
\begin{align}
\label{u_def}
\U(S; \xgen) := \log p_S(\xgen). 
\end{align}
This likelihood function serves as a natural utility measure for generative models, inspired by the fundamental principles of maximum likelihood estimation that underpin the training of many such models. The utility offers a way to measure the extent to which the data sources from $S$ are responsible for generating the content. 
It is small if the counterfactual model is unlikely to generate the same content as the deployed model, and vice versa. If the learning algorithm $\mathcal{A}$ is stochastic, we account for this randomness by taking the expectation over the algorithm. In such cases, the utility is defined as $\U(S; \xgen) := \E_{\A}[\log p_S(\xgen)]$.

In practice, the generation of the content involves prompts and human interactions, from which we can write the density conditional on an event $Q$. The utility definition becomes $\U(S; \xgen) = \log p_S(\xgen | Q)$. More generally, the utility can be defined relative to a baseline model, which, for example, is trained only on data in the public domain (that is, $S$ is the empty set $\emptyset$). The relative utility is defined as 
\begin{equation}\label{u_def2}
\U(S; \xgen) = \log \frac{p_S(\xgen | Q)}{p_{\emptyset} (\xgen | Q)}.
\end{equation}
This formulation can be viewed as the \emph{additional} information about $\xgen$, in bits, contributed by the data of $S$ beyond what is available in the public domain dataset.

\paragraph{Royalty sharing among copyright owners} The utility \eqref{u_def} or \eqref{u_def2} can be interpreted as the total compensation all members of $S$ collectively deserve for providing their data to train the generative AI model. The next step is to determine the payoff for each individual copyright owner, based on the utilities of all possible combinations of data sources. We propose using the \emph{Shapley value} \cite{shapley1953value} for this task. The Shapley value is a solution concept in cooperative game theory that offers a principled approach to distributing gains depending on the utility of every combination of players as a coalition. It is the only payoff rule that satisfies several crucial economic properties \cite{shapley1953value,roth1988introduction}, including efficiency, symmetry, dummy, and linearity. These properties are essential for ensuring fair revenue distribution among copyright owners (see supplementary materials for details).




Given the utility $\U(S)$ defined in \eqref{u_def} or \eqref{u_def2}, the Shapley value of the $i$th copyright owner is defined as
\begin{align}
&\phi_i := \frac{1}{n} \sum_{k=1}^{n} {n-1 \choose k-1}^{-1} \sum_{\substack{S \subseteq N \setminus \{i\}\\ |S|=k-1}} \left[ \U(S \cup \{i\}) - \U(S) \right].
\label{eq:shapley-formula}
\end{align}
The Shapley value remains the same regardless of whether the absolute utility \eqref{u_def} or its relative counterpart \eqref{u_def2} is being used. At a high level, it rewards a copyright owner based on the weighted average of utility changes caused by adding this contributor's data to all possible coalitions. The Shapley value is large if the addition of the data source enhances the likelihood of generating the artwork for many combinations of contributors' data in training. In particular, it equals zero if the contributor's data does not impact the likelihood of generating the content $\xgen$ for any combination.\footnote{We note that, in practice, due to learning stochasticity, utility functions are randomized, rendering the Shapley value a random variable. While previous research has demonstrated that such stochasticity can significantly affect the estimation of Shapley values when each player contributes only a single data point, this paper focuses on scenarios where each player possesses a data source \cite{wang2023data}. In such settings, the impact of learning stochasticity on Shapley value estimation is minimal.}

In our framework, the $i$th copyright owner receives a payoff proportional to the following Shapley Royalty Share (SRS) for the AI-generated content $\xgen$:
\begin{equation}\label{eq:prop}
\frac{\phi_i( \U(\cdot; \xgen) )
}{\sum_{j=1}^n \phi_j( \U(\cdot; \xgen) )},
\end{equation}
where the denominator equals the total relative utility $\U(N; \xgen)$ defined in \eqref{u_def2} due to the efficiency property of the Shapley value \cite{shapley1953value}. When $\phi_i( \U(\cdot; \xgen) )$ is negative, we replace it by zero in both the numerator and denominator of \eqref{eq:prop}. For instance, if the user pays one dollar for generating $\xgen$, a copyright owner would receive a payoff in the amount of its SRS. However, in practice, it is reasonable for the AI developers to retain a fraction of the revenue since it costs considerable resources to train the model. We defer the discussion of this point to Section \ref{sec:discuss}.

From the definition of the SRS in \eqref{eq:prop}, if a contributor's data has a relatively large Shapley value, this contributor would receive a large royalty share, and vice versa. 
As the Shapley value is a fair metric of each party's contribution to the coalition~\cite{shapley1953value}, the SRS offers a principled approach to assigning royalty shares to copyright owners. 
A related approach is called leave-one-out (LOO) score \cite{cook1980characterizations, deng2023computational}, which examines the effect of removing a single data point or source from the entire training dataset. However, it may not capture the complex interactions among various data sources. 
This shortcoming becomes especially pronounced with data duplication across various copyright owners, which is common in machine learning applications \cite{lee2022deduplicating}; see the supplementary materials for detailed discussion.

\paragraph{Computational considerations.} 
A main challenge in implementing SRS is the need to retrain the model multiple times to evaluate utility functions for different data source combinations. However, in applications with a few copyright owners, this challenge is much less pronounced. We envision this contract-based framework working best when copyrighted data is partitioned among a handful of owners, ensuring each source significantly impacts the training outcome. For very small data sources, the royalty share would be minimal and potentially noisy due to the stochastic nature of AI model training \cite{wang2023data}.

Additionally, for practical implementation in commercial AI models handling millions of daily transactions, estimating aggregated payoffs for each copyright owner, rather than calculating individual payoffs for each AI-generated content, would suffice. Computational costs can be further reduced by evaluating SRS for a small fraction of transactions and scaling to estimate overall revenue distributions.

Our work's primary contribution lies in the novel concept of using the Shapley value to address the AI copyright issue, rather than in tackling these computational challenges. The latter remains an active research area in machine learning, including Monte Carlo methods for approximating the Shapley value \cite{jia2019towards, illes2019estimation, okhrati2021multilinear,wang2021improving,burgess2021approximating,mitchell2022sampling, lin2022measuring, wang2023notegroup}, particularly useful for large coalitions of copyright owners. Another approach to alleviating computational burdens in calculating Shapley values is the In-Run Data Shapley method \cite{wang2024data}. This method significantly reduces computational overhead by computing Shapley values during the training process, rather than requiring multiple retraining runs on different data subsets. Our SRS framework lays the conceptual foundation for using Shapley values in AI copyright disputes, while acknowledging that efficient computation of these values remains an important area for future research and practical implementation.

\paragraph{Beyond copyright considerations}
Our framework not only tackles copyright disputes but also addresses scenarios where multiple entities, each holding a private dataset, seek to jointly train a generative AI model with the objective of generating revenue from its application. Though initially driven by copyright concerns, our SRS framework adapts seamlessly to these new scenarios, ensuring fair revenue sharing among private data owners. Crucially, this approach addresses potential financial disagreements and facilitates decentralized model development.

\begin{remark}[Adversarial data owners]
We assume the data owners are honest and follow the protocol of SRS. In practice, an owner might attempt to artificially inflate their contribution by including non-copyrighted images or even AI-generated images in their dataset. 
Investigating robust methods to detect and mitigate such adversarial behavior will be crucial for ensuring the long-term integrity and fairness of the royalty-sharing framework in AI-generated content. These potential adversarial settings are important future directions and are beyond the scope of the current work.
\end{remark}

\section{Experiment}
\label{sec:experiments}

We assessed our proposed framework's effectiveness in distributing royalties for AI-generated content across multiple copyright-sensitive domains: creative art, logo design, and language generation. 
We used three publicly available datasets: (1) WikiArt \cite{saleh2016large} for creative art, (2) FlickrLogo-27 \cite{kalantidis2011scalable} for logo design, and (3) Pile \cite{gao2020pile} for language generation. Detailed dataset descriptions and training algorithm settings are provided in the supplementary materials.

\subsection{Evaluation protocol} 
\label{sec:eval-protocol}

Our evaluation protocol was designed to simulate real-world scenarios where AI models are trained on diverse datasets with multiple copyright owners. For each domain, we pretrain a model on the non-copyrighted dataset, followed by fine-tuning on copyrighted dataset. We finetune on different combinations of the copyrighted datasets to compute the SRS and assess its effectiveness in attributing contributions from various sources. 


\paragraph{Creative Art (WikiArt).} We selected four disjoint subsets of paintings from four renowned artists (Van Gogn, Monet, Picasso, Rembrandt) as the copyrighted datasets. Our base model was trained on all artworks in the WikiArt dataset, excluding those belonging to the four selected artists.



\paragraph{Logo Design (FlickrLogo-27).} Similarly, we selected four disjoint subsets of logo designs from four brands (Sprite, Starbucks, Vodafone, Google). The SRS was computed using a base model trained on logo images from other brands not included in the selected four.


\paragraph{Language generation (Pile).} For the task of language generation, we use the Pile dataset.\footnote{\url{https://huggingface.co/datasets/monology/pile-uncopyrighted}.} Specifically, we first pretrain a Pythia 410M model on Pile-CC, a subset of the Pile \cite{gao2020pile} consisting of a general web corpus. We simulate the scenario where each owner of copyrighted data possesses a specific domain within the Pile, including \emph{PubMed Abstracts}, \emph{PubMed Central}, \emph{Wikipedia}, \emph{GitHub}, and \emph{EuroParl}. We then estimate the Shapley values by fine-tuning the model on different combinations of these domains.



\subsection{Results}

\begin{figure}[h]
    \centering
    \includegraphics[width=\linewidth]{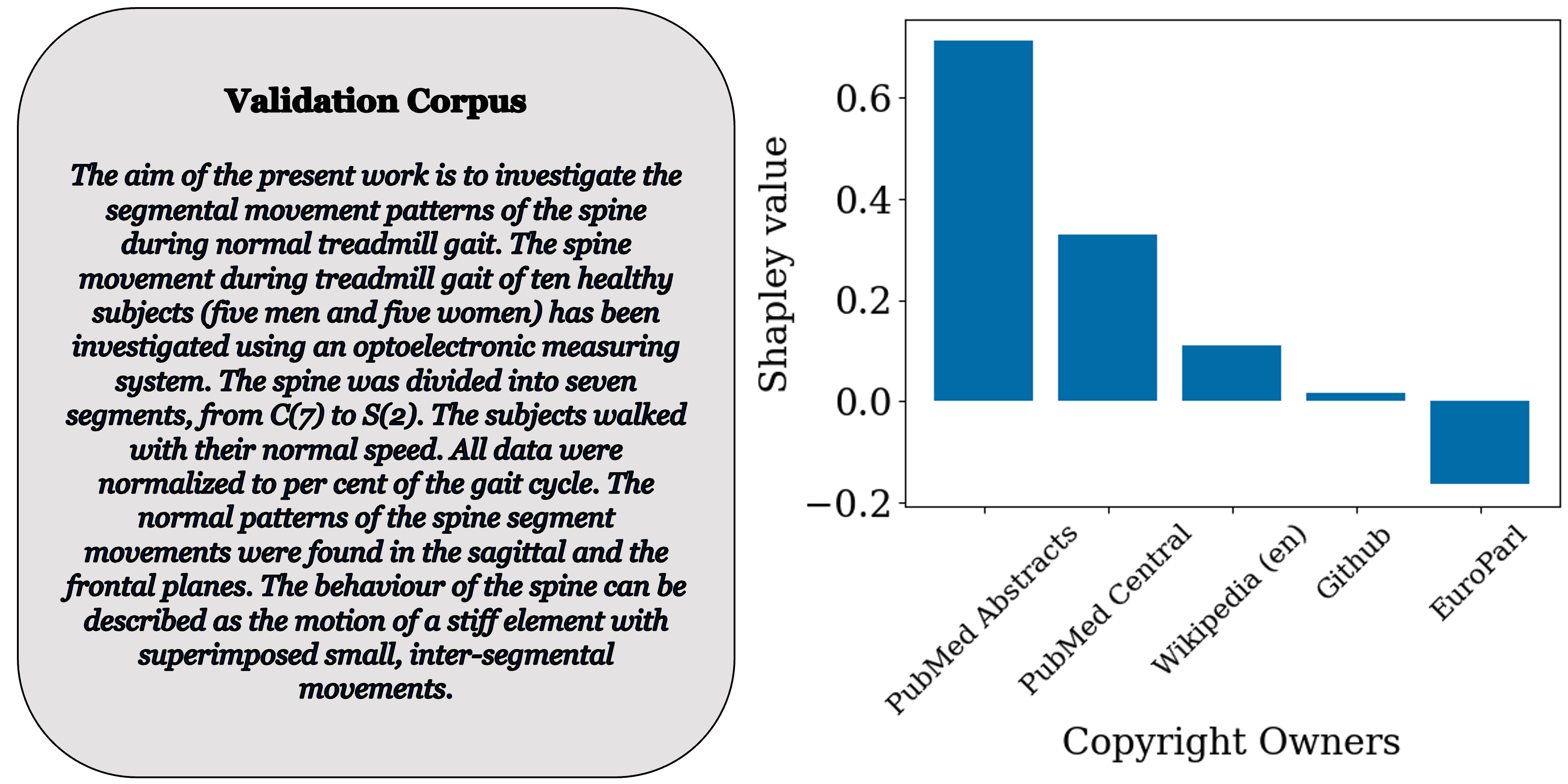}
    \caption{
    Left: The corpus from \textbf{PubMed Abstract} we use as the query example for contribution evaluation. 
    Right: The Shapley value for the copyrighted owners.
    }
    \label{fig:pubmed}
\end{figure}

\paragraph{Identifying relevant copyright owners} 
Figure~\ref{fig:wikiart} shows the computed SRS for different kinds of $\xgen$'s that are either the original or the AI-generated painting that is in the style of different artists. The results indicate that the SRS has the highest values when the $\xgen$'s closely resembles the training data source in style. This relationship underscores the SRS framework's ability to accurately attribute contributions to the creation of AI-generated images. Figure \ref{fig:pubmed} further illustrates this concept in the context of language models by displaying the Shapley values for different copyrighted subdomains of the Pile dataset in relation to generating a corpus in PubMed Abstract format. As expected, the PubMed Abstract subdomain achieves the highest Shapley value for the language model, while PubMed Central also scores highly due to domain similarity. Wikipedia, being a more general domain, shows a smaller but positive contribution to the model's output. In contrast, EuroParl, which contains a significant amount of foreign language content, exhibits a negative Shapley value for the language model, likely due to the substantial domain shift from the target medical literature domain. More results are available in the supplementary materials (Section \ref{appendix:eval-results}).

\begin{figure}[h]
    \centering
    \includegraphics[width=\linewidth]{images/combine_copyright_owners.pdf}
    \caption{
    Evaluation of the SRS using the WikiArt (upper) and \logo~datasets (lower): 
    Each row displays example target images ($\xgen$'s) for which the SRS is assessed. 
    Left: The heatmap of the SRS of copyright owners in producing the 
    original paintings from different artists (or original logo designs from different brands). 
    Right: The heatmap of the SRS of copyright owners in producing AI-generated paintings in the style of different artists (or AI-generated logo designs of different brands). 
    } 
    \label{fig:wikiart}
\end{figure}

\begin{figure}[h]
    \centering
    \includegraphics[width=\linewidth]{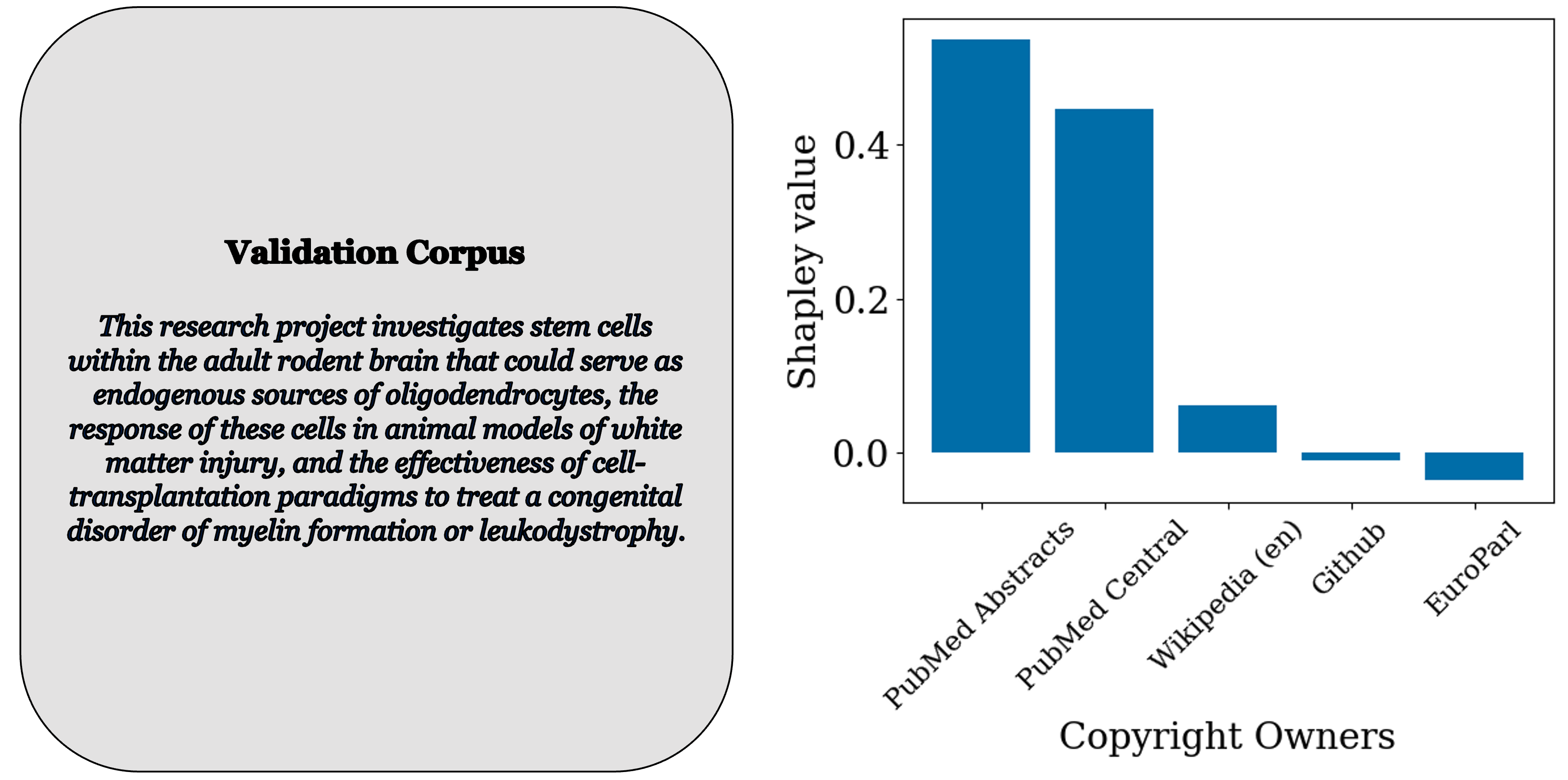}
    \caption{
    Left: The corpus from \textbf{NIH ExPorter}, a medical dataset we use as the query example for contribution evaluation. 
    Right: The Shapley value for the copyrighted owners.
    }
    \label{fig:nih-exporter}
\end{figure}

\begin{figure}[h]
    \centering
    \includegraphics[width=\linewidth]{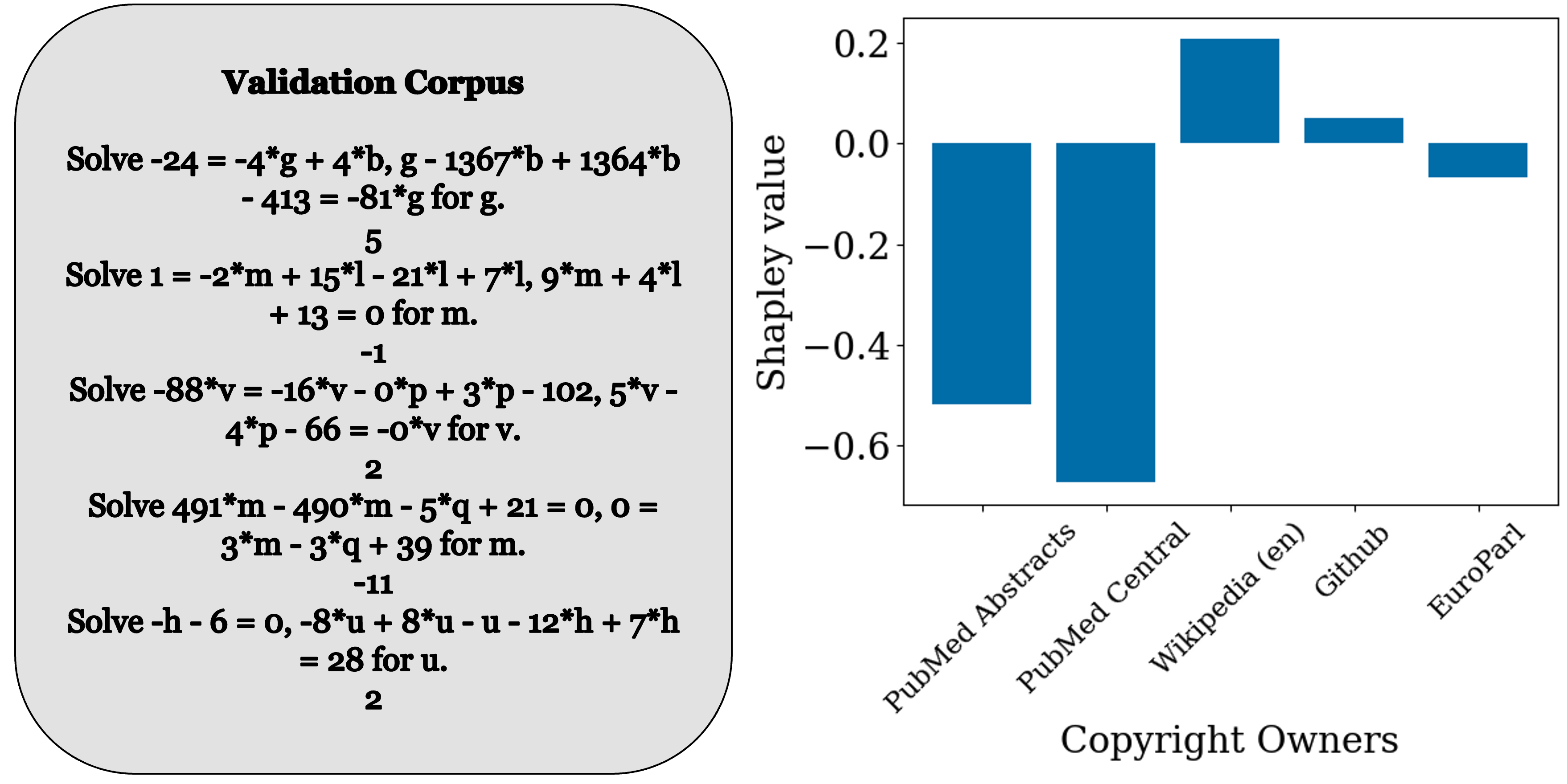}
    \caption{
    Left: The corpus from DeepMind Mathematics (\textbf{DM Mathematics}), a corpus that contains simple math questions which we use as the query example for contribution evaluation. 
    Right: The Shapley value for the copyrighted owners.
    }
    \label{fig:dmmathematics}
\end{figure}

\begin{figure}[!htp]
    \centering
    \includegraphics[width=\linewidth]{images/Mix_cp_owners-2.pdf}
    \caption{Results on the WikiArt dataset when prompting generative AI to produce a painting based on multiple copyright owners' style. 
    Left: The generated images. 
    Right: The histogram of the SRS of different copyright owners when using AI-generated images.
    } 
    \label{fig:wikiart-combine-1}
\end{figure}

\paragraph{Assessing mixed-style generation}
In Figure \ref{fig:wikiart-combine-1}, we explored the SRS distribution for prompts requesting content generation from multiple data sources. Notably, for the WikiArt dataset, prompts asked the generative model to blend styles from multiple artists. The SRS effectively recognized and rewarded the contributions of data sources integrated into the generated artworks, showcasing the framework's capability to discern and value diverse data source inputs to generate content. Figure \ref{fig:nih-exporter} illustrates the Shapley values when evaluating a corpus from NIH Exporter, a medical dataset similar to ``PubMed Abstracts'' and ``PubMed Central''. As anticipated, both ``PubMed Abstracts'' and ``PubMed Central'' achieved the highest value scores, reflecting their close domain similarity to the NIH Exporter dataset. This result further validates the SRS framework's capability to accurately attribute relevance to closely related data sources in specialized domains.


\begin{figure}[h]
    \centering
    \includegraphics[width=\linewidth]{images/Irrelevant_cp_owners.pdf}
    \caption{Results on FlickrLogo-27 Dataset when prompting generative AI for producing logos for Coca-Cola / DHL, brands whose logo images are not contained in any of the copyright owners' training set. 
    Left: The generated images. 
    Right: The histogram of the SRS of different copyright owners for the AI-generated images. 
    } 
    \label{fig:logo-irrelevant-1}
\end{figure}

\paragraph{Non-copyrighted data} 
We further explored the SRS framework's response to prompts requesting content generation from non-copyrighted data sources, as shown in Figure \ref{fig:logo-irrelevant-1}. 
In these scenarios, the SRS distribution was observed to be nearly uniform across all copyright owners. 
This outcome aligns with expectations, as the generated content lacks direct ties to any of the copyrighted data sources. This uniformity demonstrates the SRS framework's ability to avoid disproportionate revenue distribution. Figure \ref{fig:dmmathematics} presents the Shapley values for generating a corpus in DM Mathematics. Notably, all domains exhibit very low or negative Shapley values, with the exception of Wikipedia. This exception can be attributed to Wikipedia's substantial repository of math-related content, which likely contributes positively to the generation of mathematical text. This result underscores the SRS framework's sensitivity to domain-specific relevance, even within broader, general-purpose datasets.


\paragraph{Ranking of contributions via SRS} 
In many applications, it is essential to understand the hierarchy of contributions from data sources. To validate the SRS framework's capability to do so, we conducted experiments using the CIFAR100 dataset \cite{krizhevsky2009learning}, focusing on four distinct categories: Aquarium Fish, Other Fish, Aquatic Mammals, and Furniture. With ``Aquarium Fish'' images as the baseline for generation, it is natural to expect the following relevance order: Aquarium Fish $>$ Other Fish $>$ Aquatic Mammals $>$ Furniture. Figure~\ref{fig:cifar10-rank} shows that the SRS framework accurately reflects this expected ordering, demonstrating its robustness in discerning the relative significance of contributions from diverse data sources.

\begin{figure}[!htp]
    \centering
    \includegraphics[width=\linewidth]{images/cifar100-rank_cp.pdf}
    \caption{
    We divide the CIFAR100 training set into 4 disjoint groups, where each group has a different level of similarity to the target images, and then evaluate the SRS. 
    }
    \label{fig:cifar10-rank}
\end{figure}








\section{Related Work} 
\label{sec:relworks}

Recent efforts in machine learning have primarily focused on minimizing the likelihood of creating copyright-infringing content by generative AI models. One approach involves training an auxiliary generative model on non-copyrighted data and utilizing rejection sampling to reduce the likelihood of reproducing copyrighted material \cite{vyas2023provable}. However, this method is susceptible to adversarial attacks \cite{li2023probabilistic}. Alternatively, \cite{chu2023protect} suggested modifying generative models' training objectives to avoid generating outputs that closely resemble copyrighted data. Yet another technique focuses on protecting unique artistic styles by incorporating adversarial perturbations into copyrighted images for model fine-tuning~\cite{shan2023glaze}. More recently, techniques such as rewriting the input prompt \cite{he2024fantastic} and modifying outputs to be more generic \cite{chibaokabe2024tackling} have also been proposed.

The Shapley value has been suggested as a means to fairly distribute revenue in traditional sectors such as royalty agreements between music copyright holders and radio broadcasters \cite{watt2014fair}. More recently, it has been used for data valuation where the utility function is the prediction accuracy of the machine learning model \cite{ghorbani2019data, jia2019towards}. This differs from our SRS framework, which uses the log-likelihood as the utility since there is no such thing as prediction accuracy for generative models.

There are a few techniques that have been extended to the generative models. A simple approach utilizes similarity scores between training data and generated content as a valuation metric \cite{yang2023matching}. Another commonly used approach is the leave-one-out (LOO) score or its approximations. For example, \cite{georgiev2023journey} extends the TRAK framework \cite{park2023trak} to generative models, and \cite{zheng2023intriguing} further introduced empirical approaches to improving the performance of \cite{georgiev2023journey}. 
Notably, \cite{deng2023computational} proposed a revenue-sharing mechanism for AI-generated music based on TRAK, which is closely related to our work. However, the LOO scores neglect the high-order training data interactions, which may result in undesirable attribution scores (see Section \ref{appendix:loo-vs-shapley} in the supplementary materials for detailed discussion).

\section{Discussion}
\label{sec:discuss}

The recent rise of generative AI has profoundly challenged traditional copyright laws, driven by its powerful generating capabilities. This is compounded by the intricacies in the interpretation of copyrights for AI-generated content as well as the black-box nature of large AI systems. We have addressed these issues from an economic standpoint by developing a royalty sharing model that permits training on copyrighted data in exchange for revenue distribution among copyright owners. This fosters mutually beneficial cooperation between the AI developers and copyright owners. Our framework has several economic underpinnings that render it fair and interpretable. We demonstrate the effectiveness and feasibility of this framework through numerical experiments.

Our study, however, has limitations and opens avenues for future investigation. One concern is potential strategic behaviors, such as copyright owners merging or splitting their data to maximize their royalty share. The SRS could be manipulated by a malicious copyright owner creating multiple copies of their data. While replication-robust solution concepts have been explored \cite{han2020replication}, they focused on the impact on Shapley values rather than ratios under replication. Developing a mechanism robust against such manipulation is an important direction for future work. Another open question is handling copyrighted data when owners are unable or unwilling to negotiate agreements, particularly with numerous owners each having small datasets. In such cases, our approach could be combined with methods for generating lawful content \cite{vyas2023provable}. Enhancing our model to determine appropriate revenue division between copyright owners and AI developers, acknowledging the critical role of computational resources, algorithm design, and engineering expertise in developing high-performance AI models, is another avenue for research. We have made preliminary progress toward this by adapting the concept of permission structure from cooperative game theory \cite{gilles1992games} to model the scenario where the AI developers and copyright owners jointly train a generative AI; see the supplementary materials for details.

From a methodological perspective, a crucial aspect warranting future research is the use of Shapley value ratios for revenue distribution. The key challenge with directly using the Shapley value lies in the unknown total revenue for any coalition of copyright owners' data. The log-likelihood ratio \eqref{u_def2} serves as a surrogate for this unknown quantity. However, the efficiency property of the Shapley value \cite{shapley1953value}, which ensures the sum of Shapley values equals the grand coalition's utility, loses meaning when considering ratios. In this light, semivalues \cite{dubey1981value}, which are a generalization of the Shapley value that drop the efficiency axiom, could provide a viable alternative. Future work could aim to establish axiomatic justifications to identify the most suitable solution concepts within the semivalue class for royalty distribution in this context.

\input{acknowledge}

\bibliographystyle{plain}
\bibliography{ref}


\newpage
\onecolumn

\appendix

\section{Supplementary Materials}
\label{sec:discussion}

Recall the setting where we have $n$ copyright owners where each copyright owner $i \in N = \{1, \ldots, n\}$ owns the copyright of training data $D^{(i)}$. 
The utility function $\U$ is defined as 
\begin{align*}
\U(S; \xgen) = \log p_S(\xgen),
\end{align*}
where $p_S(\cdot)$ denotes the probability density function of the model trained on $\cup_{i \in S} \Di$. 
The Shapley value of the copyright owner $i$ for generating $\xgen$ is 
\begin{align*}
&\phi_i = \frac{1}{n} \sum_{k=1}^{n} {n-1 \choose k-1}^{-1} \sum_{\substack{S \subseteq N \setminus \{i\}\\ |S|=k-1}} \left[ \U(S \cup \{i\}) - \U(S) \right].
\end{align*}

\label{appendix:SV-background}

The Shapley value is a concept from cooperative game theory and provides a principled approach to fairly distribute the total gains (or costs) among coalition participants based on their individual contributions. 
The theoretical foundation of the Shapley value is established through four axioms introduced by Lloyd Shapley in his 1953 paper \cite{shapley1953value}. These axioms delineate criteria for an equitable and logical distribution of payoffs, where the Shapley value is the \emph{unique} solution concept that satisfies all of them. Recall that $N = \{1, \ldots, n\}$ the set of players. 

\begin{itemize}
    \item Dummy player: if $\U\left(S \cup i\right)=\U(S)+c$ for all $S \subseteq N \setminus \{i\}$ and a scalar $c$, then $\phi_i = c$. 
    \item Symmetry: if $\U(S \cup i) = \U(S \cup j)$ for all $S \subseteq N \setminus \{i, j\}$, then $\phi_i=\phi_j$. 
    \item Linearity: for utility functions $\U_1, \U_2$ and any scalar values $\alpha_1, \alpha_2$, 
    \[
    \phi_i\left( \alpha_{1} \U_{1}+\alpha_{2} \U_{2}\right)=\alpha_{1} \phi_i \left( \U_{1}\right)+
    \alpha_{2} \phi_i \left( \U_{2}\right).
    \]
    \item Efficiency: for every $\U, \sum_{i \in N} \phi_i = \U(N)$. 
\end{itemize}

In plain words, the \emph{efficiency} axiom requires the total value to be distributed among individuals. The \emph{symmetry} and \emph{null player} axiom refer to ``same contribution, same value'' and ``no contribution, no value'', respectively. The \emph{linearity} axiom requires the value scores add up when utility functions add up. 
These principles lay the groundwork for a revenue distribution method that ensures that every participant receives a share of the total value that reflects their contribution to the coalition. The Shapley value's unique ability to satisfy these conditions makes it a powerful tool for analyzing cooperative scenarios and allocating resources or costs in a manner that is widely considered fair.

\subsection{Efficient SRS Estimation for High-volume Transactions}
\label{appendix:SRS-multiple}

Considering the potential for millions of daily uses and transactions of a commercial AI model, computing the SRS for each individual transaction may not be feasible due to the high computational cost. 
To address this challenge, a practical solution is to estimate the average SRS for each copyright owner based on a subset of transactions. For any given transaction involving AI-generated content $\xgen$, let $p_{\xgen}$ represent the payment by the user. The share of this payment received by copyright owner $i$ is determined by:
$$
p_{\xgen} \cdot \SRS_{\xgen},
$$
where $\SRS_{\xgen} := \frac{\phi_i( \U(\cdot; \xgen) )}{\sum_{j=1}^n \phi_j( \U(\cdot; \xgen) )}$. Ideally, the goal is to compute the cumulative payment distribution over all of the daily transactions:
$$
\sum_{\xgen} p_{\xgen} \cdot \SRS_{\xgen}.
$$
Given the computational cost of SRS calculations, a Monte Carlo method can be applied to randomly sample a batch of $\xgen$ transactions and estimate $\E_{\xgen} [\SRS_{\xgen}]$, and then multiply it by the total amount of the income in a day $\sum_{\xgen} p_{\xgen}$. 
This estimation remains unbiased as long as $\SRS_{\xgen}$ and $p_{\xgen}$ are independent, allowing for:
$$
\E_{\xgen} \left[ p_{\xgen} \SRS_{\xgen} \right] \approx \E_{\xgen} [p_{\xgen}] \E_{\xgen} [\SRS_{\xgen}].
$$

We note that, in practice, $p_{\xgen}$ is often a constant number as the charges for different user queries are usually of the same rate. In this case, $p_{\xgen}$ and $\SRS_{\xgen}$ are clearly independent.

\subsection{The SRS of AI Developers}
\label{appendix:AI-Company}

Recall that within our framework, each copyright owner, denoted by $i$, is awarded a payoff directly proportional to their SRS for AI-generated content $\xgen$:
\begin{equation}
\frac{\phi_i( \U(\cdot; \xgen) )}{\sum_{j=1}^n \phi_j( \U(\cdot; \xgen) )}.
\end{equation}
However, in real-world applications, the AI developing company decides on a collective share, $\ShareData$, ranging from 0 to 1, designated for distribution among the copyright owners. The remaining share of the royalty, quantified as $1-\ShareData$, is retained by the company. Consequently, the adjusted royalty share for each copyright owner concerning AI-generated content $\xgen$ is calculated as:
$$
\ShareData \cdot \frac{\phi_i( \U(\cdot; \xgen) )}{\sum_{j=1}^n \phi_j( \U(\cdot; \xgen) )}.
$$

While the determination of $\ShareData$ typically falls to the discretion of the AI developer responsible for training the generative models, we propose a principled approach grounded in cooperative game theory to decide $\ShareData$. 
By including the AI developer as a special player in the SRS framework, we can model the scenario as a game with a ``permission structure'' \cite{gilles1992games}. Specifically, the permission structure, which is usually represented as a directed graph where nodes symbolize players and directed edges signify permissions, determines valid coalitions. A coalition is considered valid only if every member has the requisite permissions, either inherently or through other coalition members.

In our context, incorporating the AI developer as an additional player expands the player set to $\playerset := N \cup \{\aicomp\}$. 
The presence of the AI developer is indispensable for model training, making it impossible to achieve any utility without their involvement. Consequently, the utility function for $\playerset$ is adjusted to $\U^\star(S) = \U(S \setminus \{\aicomp\})$ when $\aicomp$ is included in $S$, and $\U^\star(S) = 0$ for coalitions excluding $\aicomp$. 
To reasonably define $\ShareData$, we calculate the Shapley value (and consequently the SRS) of the AI developer within this modified game structure, offering a quantifiable metric of the AI developer's contribution.

\section{Comparison Between Leave-one-out and the Shapley Value}
\label{appendix:loo-vs-shapley}

The leave-one-out (LOO) score \cite{cook1980characterizations} is a simple, straightforward method for assessing the contribution of data sources. Specifically, the LOO score of a copyright owner is calculated as the model performance change when the data source belonging to the copyright owner is excluded from the full training set:
$$
\phi_i^{\loo} := \U(N) - \U(N \setminus \{i\}).
$$
Although intuitive for evaluating the impact of individual data sources, the LOO score has limitations. It solely examines the consequence of removing a data source from the entire dataset. This approach might not accurately reflect the significance of a data point due to potential complex interactions among data sources. 
Duplicated data points are prevalent across many widely-used machine learning datasets \cite{lee2022deduplicating}. 
Consider two copyright owners $i$ and $j$ having nearly identical data. The removal of either from the dataset would likely result in minimal change to the model's content generation likelihood, rendering both LOO scores close to zero. This scenario could unjustly allocate no royalty share to either contributor, despite their datasets' crucial role in model performance. Moreover, in situations with numerous data sources, the LOO score might diminish to near zero, failing to recognize the nuanced contributions of individual sources.

In contrast, the Shapley value method accounts for the incremental impact of incorporating a data source alongside all possible combinations of other sources. This comprehensive approach effectively captures the intricate dynamics among data sources, offering a more accurate and fair assessment of each contribution.

\section{Experimental Settings}
\label{appendix:setting}

\paragraph{Datasets}
Our study focuses on art painting and logo design, two domains where copyright plays a pivotal role in safeguarding the integrity and commercial value of creative outputs. 
For art paintings, our research employs the WikiArt dataset \cite{saleh2016large}, which comprises approximately 80,000 artworks spanning the last 400 years. This collection features pieces from over 1,000 renowned artists with a wide variety of styles and genres. For logo design, we use FlickrLogo-27 dataset \cite{kalantidis2011scalable}, which consists of images from 27 distinct logo classes or brands, sourced from Flickr.

\paragraph{Model architectures \& training details} 
The generative models used in our experiment follow from the recent advancements in high-resolution image synthesis utilizing latent diffusion models \cite{rombach2022high}.\footnote{Part of the codebase is adapted from \url{https://github.com/artem-gorodetskii/WikiArt-Latent-Diffusion} and \url{https://github.com/VSehwag/minimal-diffusion}.}
All images were cropped and resized to $512 \times 512$ resolution. 
As training a new generative model from scratch would be prohibitively costly, we use LoRA \cite{hu2021lora}, an efficient fine-tuning method that enables us to scale up the models being used in a tractable manner.\footnote{
The implementation is adapted from \url{https://huggingface.co/blog/lora}. 
} 
The model used for fine-tuning is Stable Diffusion V1-4\footnote{\url{https://huggingface.co/CompVis/stable-diffusion-v1-4}.}, which is a latent image diffusion model trained on LAION2B-en\footnote{\url{https://huggingface.co/datasets/laion/laion2B-en}.}. 
For the WikiArt dataset, we train each model on painting images from the same artist, with the text prompt ``A painting in the style of \texttt{[artist name]}.'' 
Similarly, for the FlickrLogo-27 dataset, we train each model on logo images from the same artist, with the text prompt ``A logo by \texttt{[company name]}.'' 
For each dataset, the model is fine-tuned with an initial learning rate $10^{-4}$, 10 epochs, and batch size $4$. 
We train the diffusion models on A100 GPU cluster. 
For all the SRS results in the maintext, they are being averaged over 20 $\xgen$'s that are randomly selected/generated according to the specifications in the maintext. 

\paragraph{Calculating log-likelihood of AI-generated content}
The diffusion models have two main processes: the forward (noise-adding) process and the reverse (noise-removing) process. The \emph{forward process} is a Markov chain that gradually adds Gaussian noise to the data over a series of steps. If we represent the original data as $x_0$, the process of adding noise can be expressed as:
$$
x_t = \sqrt{\alpha_t} x_{t-1} + \sqrt{1 - \alpha_t} \epsilon,
$$
where $x_t$ is the data at step $t$, $\epsilon$ is a sample from a standard Gaussian distribution $\mathcal{N}(0, I)$, $\alpha_t$ is a variance schedule that determines how much noise to add at each step, and $t$ ranges from 0 to $T$, with $T$ being the total number of diffusion steps, and $x_T$ being almost entirely noise. 
The \emph{reverse process} aims to learn the distribution of the original data by starting from noise and progressively removing it. This can be modeled as:
$$
p(x_{t-1} | x_t) = \mathcal{N}(x_{t-1}; \mu_{\theta}(x_t, t), \Sigma_{\theta}(x_t, t)),
$$
where $\mu_{\theta}(x_t, t)$ and $\Sigma_{\theta}(x_t, t)$ are the mean and covariance of the Gaussian distribution at step $t$, learned by the model with parameters $\theta$.\footnote{
We note that for text-to-image diffusion models, $\mu_\theta(\cdot)$ and $\Sigma_{\theta}(\cdot)$ will also depend on the input text encoding, but the core mathematical representation remains the same.
} 
The model is trained to minimize the difference between the noisy data and its prediction of the denoised data at each step. This can be formalized as minimizing a loss function, e.g., the mean squared error (MSE), between the original data and the reconstructed data:
$$
L(\theta) = \mathbb{E}_{t, x_0, \epsilon}\left[\|\epsilon - \epsilon_{\theta}(x_t, t)\|^2\right],
$$
where $\epsilon_{\theta}(x_t, t)$ is the model's prediction of the noise $\epsilon$ added at step $t$, and the expectation $\mathbb{E}$ is over different noise levels $t$, the original data $x_0$, and the noise $\epsilon$. 
To generate new data, we start with a sample from the noise distribution $x_T \sim \mathcal{N}(0, I)$ and iteratively apply the reverse process to obtain $x_{T-1}, x_{T-2}, ..., x_0$, with $x_0$ being the final generated sample.

In order to estimate the density of a diffusion model $p_{\theta}(\xgen)$ on a generated sample $x_0 := \xgen$, note that 
\begin{align}
p_{\theta}(\xgen)
&= \int p_{\theta}(\xgen, x_{1:T}) d x_{1:T} \nonumber \\
&= \int p_{\theta}(\xgen|x_{1:T}) p_{\theta}(x_{1:T}) d x_{1:T} \nonumber \\
&= \E_{x_{1:T}} \left[ p_{\theta}(\xgen|x_{1:T}) \right] \nonumber \\
&= \E_{x_{1:T}} \left[ p_{\theta}(\xgen |x_{1}) \right], \label{eq:MCsample}
\end{align}
where $x_{1:T} := (x_1, \ldots, x_T)$. 
Since 
$
p_{\theta}(\xgen |x_{1}) = \mathcal{N}(\xgen ; \mu_{\theta}(x_1, 1), \Sigma_{\theta}(x_1, 1))
$ whose probability density can be efficiently computed, we can use Monte Carlo technique to estimate $p_{\theta}(\xgen)$ based on Equation (\ref{eq:MCsample}). 
For all experiments we show in the paper, we use 20 random samples of $x_T \sim \mathcal{N}(0, I)$, apply the reverse process to obtain random samples of $x_{1:T}$ and use the sample average of $p_{\theta}(\xgen |x_{1})$ as the estimation for $p_{\theta}(\xgen)$.

\subsection{The SRS of AI developers}
\label{appendix:eval-AIdev}

In supplementary materials Section \ref{appendix:AI-Company}, we explore the SRS when the AI developer is considered a special player within a game characterized by a permission structure. 
Here in Figure \ref{fig:wikiart-comp}, we empirically evaluate the SRS in this setting. 
Specifically, Figure \ref{fig:wikiart-comp} (a) shows the result of SRS for an AI-generated painting in Van Gogh's style, and Figure \ref{fig:wikiart-comp} (b) shows the result of SRS for an AI-generated logo for Sprite. Both figures show that the AI developer achieves a markedly higher SRS compared to training data contributors. This observation aligns with the intuitive understanding that the AI developer's contribution is foundational; without their computational input and expertise, it would be infeasible to generate any valuable content.

\begin{figure}[h]
    \centering
    \includegraphics[width=0.8\linewidth]{images/aicompany_cp.pdf}
    \caption{
    (a) 
    The SRS results on paintings that are generated by prompting ``paint in Van Gogh's style.'' 
    (b)
    The SRS results on logo designs that are generated by prompting ``design a logo similar to Sprite.'' 
    }
    \label{fig:wikiart-comp}
\end{figure}

\clearpage

\subsection{Additional results}
\label{appendix:eval-results}

\begin{figure}[!htp]
    \centering
    \includegraphics[width=\linewidth]{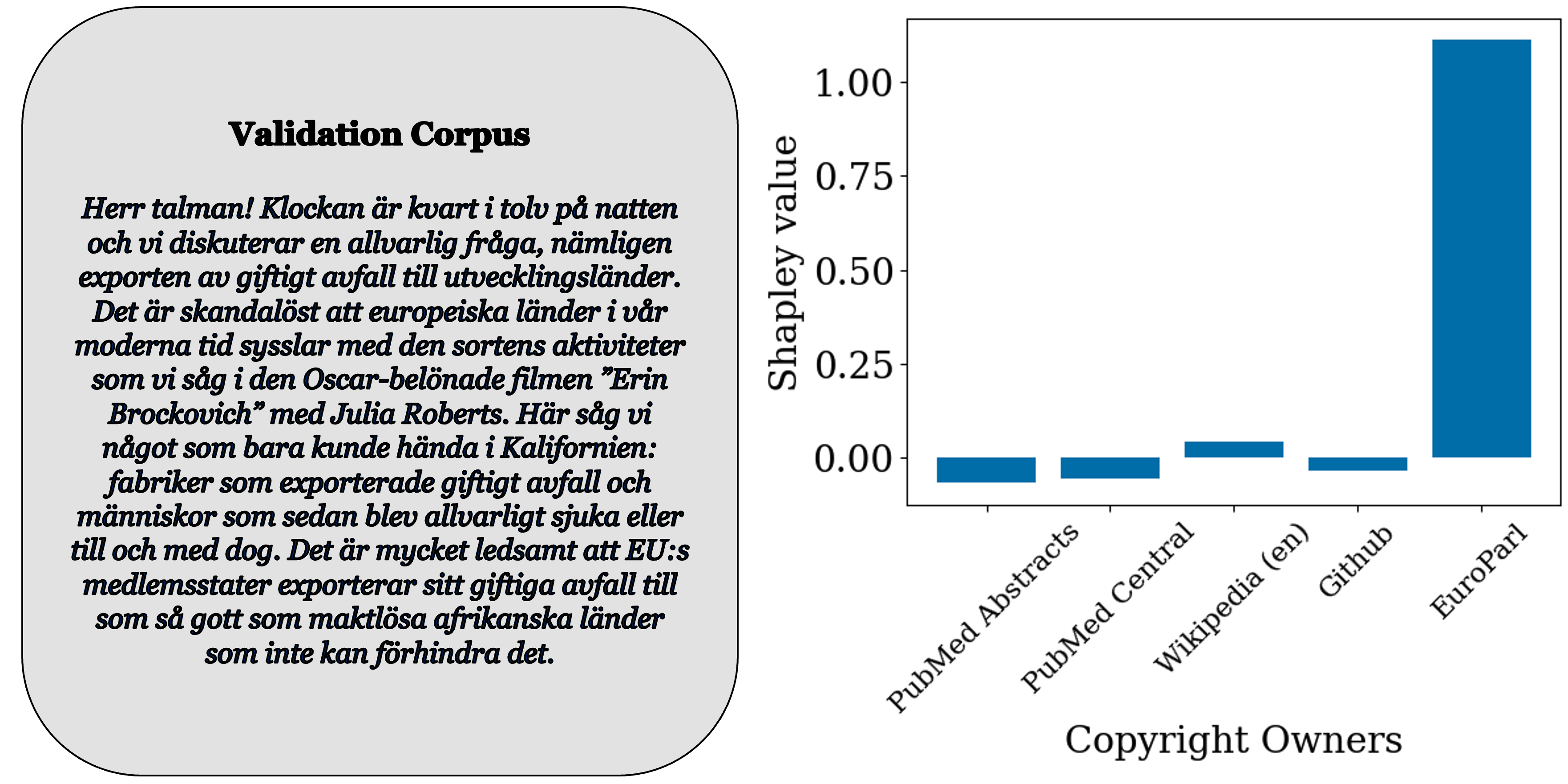}
    \caption{    
    Left: The corpus from \textbf{EuroParl}, a multilingual dataset that contains European languages which we use as the query example for contribution evaluation. 
    Right: The Shapley value for the copyrighted owners.
    }
    \label{fig:europarl}
\end{figure}

\begin{figure}[!htp]
    \centering
    \includegraphics[width=\linewidth]{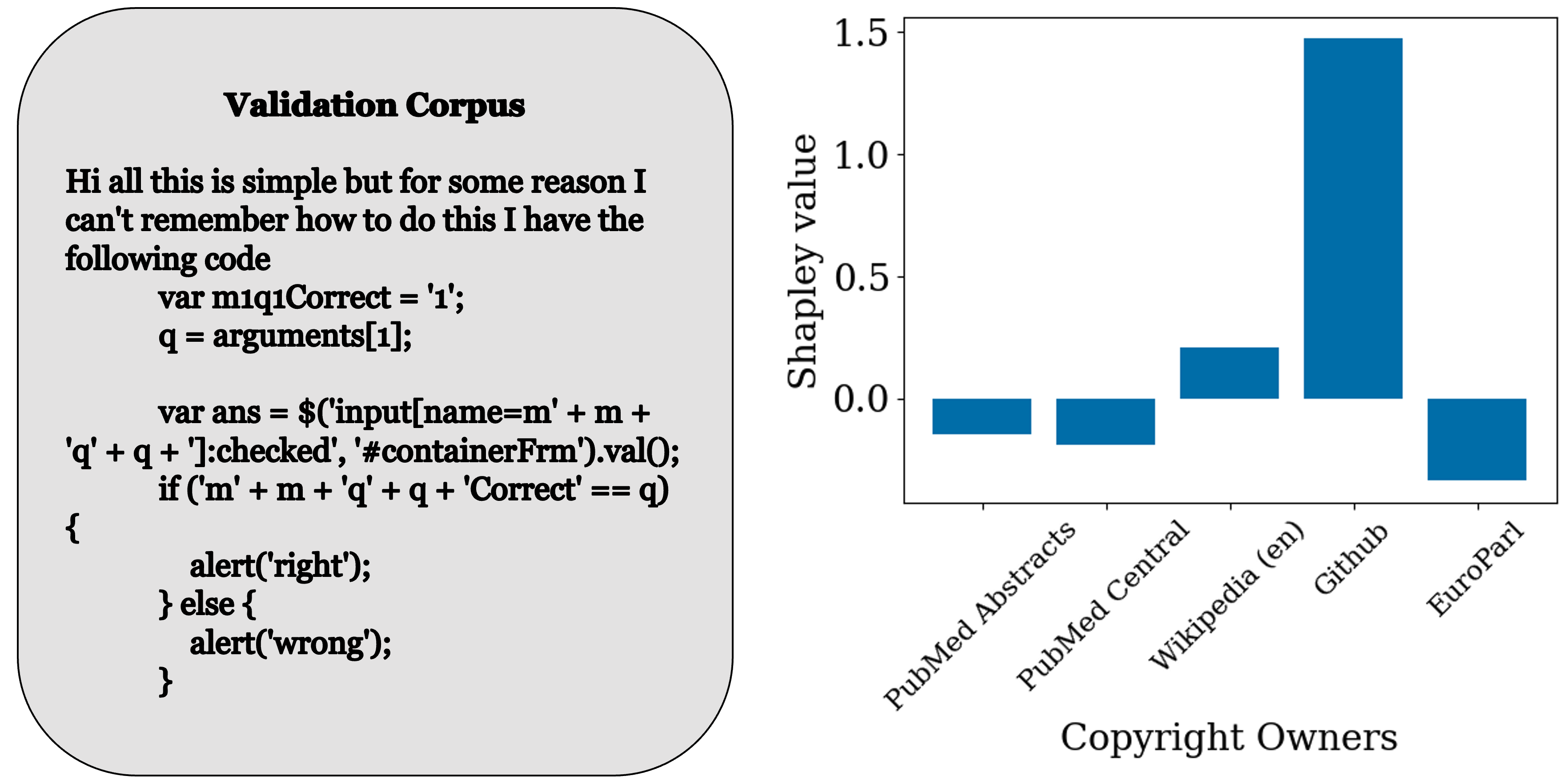}
    \caption{
    Left: The corpus from \textbf{Stack Exchange} Data Dump, which contains an anonymized set of all user-contributed content on the Stack Exchange network. 
    Right: The Shapley value for the copyrighted owners. 
    }
    \label{fig:stackexchange}
\end{figure}

\paragraph{Identifying relevant copyright owners} 
Figure \ref{fig:europarl} presents the valuation results for a validation corpus derived from EuroParl. In this case, we observe that the copyright owner of the EuroParl dataset achieves a notably high Shapley value, while all other copyright owners' Shapley values are close to zero. This stark contrast can be attributed to the fact that the EuroParl corpus contains significant non-English content, whereas the other datasets are predominantly in English.
Figure \ref{fig:stackexchange} illustrates the valuation results for a validation corpus sourced from StackExchange, which is known to contain a substantial amount of code-related data. In this scenario, we find that the copyright owner of the Github dataset attains the highest Shapley value. This outcome is logical given that the Github dataset also comprises a large volume of code-related content, aligning closely with the nature of the StackExchange corpus. 
These results further demonstrate the SRS framework's ability to accurately identify and valuate the most relevant data sources for different types of generated content, even when dealing with specialized or multilingual datasets.

\paragraph{Copyright owners with multiple datasets}
In practice, a copyright owner may possess data from multiple sources. To simulate this scenario, we present in Figure \ref{fig:domain-mixture} a case where one copyright owner holds a mixed dataset comprising EuroParl, FreeLaw, and NIH ExPorter. We evaluate this setup using a validation corpus sourced from PubMed Abstracts, which is known to contain a substantial amount of medical-related data. The results reveal that this mixed-dataset owner achieves a significantly higher Shapley value compared to an owner possessing only the EuroParl dataset (as shown in Figure \ref{fig:pubmed}). This marked increase can be attributed to the inclusion of NIH ExPorter in the mixed dataset, which also contains a significant amount of medical data.
This demonstrates that the Shapley value framework can effectively capture the increased relevance and contribution of a copyright owner when they possess diverse data sources, particularly when these sources align with the domain of the generated content.

\begin{figure}[!htp]
    \centering
    \includegraphics[width=\linewidth]{images/domain_mixture.pdf}
    \caption{
    Left: The corpus from \textbf{PubMed Abstract} we use as the query example for contribution evaluation. 
    Right: The Shapley value for the copyrighted owners, where the owner `Mixture' contains an equal amount of tokens from \textbf{EuroParl}, \textbf{FreeLaw} and \textbf{NIH ExPorter}. 
    }
    \label{fig:domain-mixture}
\end{figure}

\end{document}

%% file: macro.tex
\newif\iffinal

\iffinal
    \newcommand{\tianhao}[1]{}
    \newcommand{\ruoxi}[1]{}
    
    \newcommand{\boaz}[1]{}
    \newcommand{\wjs}[1]{}
    \newcommand{\hiroaki}[1]{}
\else
    \newcommand{\tianhao}[1]{{\bf \textcolor{purple}{[Tianhao: #1]}}}
    \newcommand{\ruoxi}[1]{{\bf \textcolor{BrickRed}{[Ruoxi:#1]}}}

    \newcommand{\boaz}[1]{{\bf {\textcolor{blue}{[Boaz :#1]}}}}
    \definecolor{wjs}{RGB}{200,0,50}
    \newcommand{\wjs}[1]{\textcolor{wjs}{[Weijie: #1]}}
    \newcommand{\hiroaki}[1]{{\color{orange} [{\bf Hiroaki:} #1]}}
\fi

%% file: acknowledge.tex
\subsection*{Acknowledgements}

JTW and ZD conducted this work as independent researchers. 
HCO, BB, and WJS were supported in part by NSF grant DMS-2310679, Wharton AI for Business, a Simons Investigator Fellowship, the Simons Foundation Math+X Grant to the University of Pennsylvania, NSF grant DMS-2134157, DARPA grant W911NF2010021, and DOE grant DE-SC0022199. We are grateful to Peter Henderson for providing helpful feedback on an early version of this paper.

%% file: arxiv.bbl
\begin{thebibliography}{10}

\bibitem{brynjolfsson2023generative}
Erik Brynjolfsson, Danielle Li, and Lindsey~R Raymond.
\newblock Generative ai at work.
\newblock Technical report, National Bureau of Economic Research, 2023.

\bibitem{bubeck2023sparks}
S{\'e}bastien Bubeck, Varun Chandrasekaran, Ronen Eldan, Johannes Gehrke, Eric
  Horvitz, Ece Kamar, Peter Lee, Yin~Tat Lee, Yuanzhi Li, Scott Lundberg,
  et~al.
\newblock Sparks of artificial general intelligence: Early experiments with
  gpt-4.
\newblock {\em arXiv preprint arXiv:2303.12712}, 2023.

\bibitem{burgess2021approximating}
Mark~Alexander Burgess and Archie~C Chapman.
\newblock Approximating the shapley value using stratified empirical bernstein
  sampling.
\newblock In {\em IJCAI}, pages 73--81, 2021.

\bibitem{chibaokabe2024tackling}
Hiroaki Chiba-Okabe and Weijie~J. Su.
\newblock Tackling genai copyright issues: Originality estimation and
  genericization.
\newblock {\em arXiv preprint arxiv:2406.03341}, 2024.

\bibitem{chu2023protect}
Timothy Chu, Zhao Song, and Chiwun Yang.
\newblock How to protect copyright data in optimization of large language
  models?
\newblock {\em arXiv preprint arXiv:2308.12247}, 2023.

\bibitem{cook1980characterizations}
R~Dennis Cook and Sanford Weisberg.
\newblock Characterizations of an empirical influence function for detecting
  influential cases in regression.
\newblock {\em Technometrics}, 22(4):495--508, 1980.

\bibitem{deng2023computational}
Junwei Deng and Jiaqi Ma.
\newblock Computational copyright: Towards a royalty model for ai music
  generation platforms.
\newblock {\em arXiv preprint arXiv:2312.06646}, 2023.

\bibitem{dubey1981value}
Pradeep Dubey, Abraham Neyman, and Robert~James Weber.
\newblock Value theory without efficiency.
\newblock {\em Mathematics of Operations Research}, 6(1):122--128, 1981.

\bibitem{gao2020pile}
Leo Gao, Stella Biderman, Sid Black, Laurence Golding, Travis Hoppe, Charles
  Foster, Jason Phang, Horace He, Anish Thite, Noa Nabeshima, et~al.
\newblock The pile: An 800gb dataset of diverse text for language modeling.
\newblock {\em arXiv preprint arXiv:2101.00027}, 2020.

\bibitem{georgiev2023journey}
Kristian Georgiev, Joshua Vendrow, Hadi Salman, Sung~Min Park, and Aleksander
  Madry.
\newblock The journey, not the destination: How data guides diffusion models.
\newblock 2023.

\bibitem{ghorbani2019data}
Amirata Ghorbani and James Zou.
\newblock Data shapley: Equitable valuation of data for machine learning.
\newblock In {\em International Conference on Machine Learning}, pages
  2242--2251. PMLR, 2019.

\bibitem{gilles1992games}
Robert~P Gilles, Guillermo Owen, and Rene van~den Brink.
\newblock Games with permission structures: the conjunctive approach.
\newblock {\em International Journal of Game Theory}, 20(3):277--293, 1992.

\bibitem{grynbaum2023times}
Michael~M Grynbaum and Ryan Mac.
\newblock The times sues openai and microsoft over ai use of copyrighted work.
\newblock {\em The New York Times}, 27, 2023.

\bibitem{han2020replication}
Dongge Han, Michael Wooldridge, Alex Rogers, Shruti Tople, Olga Ohrimenko, and
  Sebastian Tschiatschek.
\newblock Replication-robust payoff-allocation for machine learning data
  markets.
\newblock {\em arXiv preprint arXiv:2006.14583}, 2020.

\bibitem{he2024fantastic}
Luxi He, Yangsibo Huang, Weijia Shi, Tinghao Xie, Haotian Liu, Yue Wang, Luke
  Zettlemoyer, Chiyuan Zhang, Danqi Chen, and Peter Henderson.
\newblock Fantastic copyrighted beasts and how (not) to generate them.
\newblock {\em arXiv preprint arxiv:2406.14526}, 2024.

\bibitem{henderson2023foundation}
Peter Henderson, Xuechen Li, Dan Jurafsky, Tatsunori Hashimoto, Mark~A. Lemley,
  and Percy Liang.
\newblock Foundation models and fair use.
\newblock {\em arXiv preprint arXiv:2303.15715}, 2023.

\bibitem{hu2021lora}
Edward~J Hu, Phillip Wallis, Zeyuan Allen-Zhu, Yuanzhi Li, Shean Wang, Lu~Wang,
  Weizhu Chen, et~al.
\newblock Lora: Low-rank adaptation of large language models.
\newblock In {\em International Conference on Learning Representations}, 2021.

\bibitem{illes2019estimation}
Ferenc Ill{\'e}s and P{\'e}ter Ker{\'e}nyi.
\newblock Estimation of the shapley value by ergodic sampling.
\newblock {\em arXiv preprint arXiv:1906.05224}, 2019.

\bibitem{jia2019towards}
Ruoxi Jia, David Dao, Boxin Wang, Frances~Ann Hubis, Nick Hynes, Nezihe~Merve
  G{\"u}rel, Bo~Li, Ce~Zhang, Dawn Song, and Costas~J Spanos.
\newblock Towards efficient data valuation based on the shapley value.
\newblock In {\em The 22nd International Conference on Artificial Intelligence
  and Statistics}, pages 1167--1176. PMLR, 2019.

\bibitem{kalantidis2011scalable}
Yannis Kalantidis, Lluis~Garcia Pueyo, Michele Trevisiol, Roelof van Zwol, and
  Yannis Avrithis.
\newblock Scalable triangulation-based logo recognition.
\newblock In {\em Proceedings of the 1st ACM international conference on
  multimedia retrieval}, pages 1--7, 2011.

\bibitem{krizhevsky2009learning}
Alex Krizhevsky, Geoffrey Hinton, et~al.
\newblock Learning multiple layers of features from tiny images.
\newblock 2009.

\bibitem{lee2023talkin}
Katherine Lee, A~Feder Cooper, and James Grimmelmann.
\newblock Talkin''bout ai generation: Copyright and the generative-ai supply
  chain.
\newblock {\em arXiv preprint arXiv:2309.08133}, 2023.

\bibitem{lee2022deduplicating}
Katherine Lee, Daphne Ippolito, Andrew Nystrom, Chiyuan Zhang, Douglas Eck,
  Chris Callison-Burch, and Nicholas Carlini.
\newblock Deduplicating training data makes language models better.
\newblock In {\em Proceedings of the 60th Annual Meeting of the Association for
  Computational Linguistics (Volume 1: Long Papers)}, pages 8424--8445, 2022.

\bibitem{lei2023pro}
Xiaochang Lei.
\newblock Pro-rata vs user-centric in the music streaming industry.
\newblock {\em Economics Letters}, 226:111111, 2023.

\bibitem{li2023probabilistic}
Xiang Li, Qianli Shen, and Kenji Kawaguchi.
\newblock Probabilistic copyright protection can fail for text-to-image
  generative models.
\newblock {\em arXiv preprint arXiv:2312.00057}, 2023.

\bibitem{lin2022measuring}
Jinkun Lin, Anqi Zhang, Mathias L{\'e}cuyer, Jinyang Li, Aurojit Panda, and
  Siddhartha Sen.
\newblock Measuring the effect of training data on deep learning predictions
  via randomized experiments.
\newblock In {\em International Conference on Machine Learning}, pages
  13468--13504. PMLR, 2022.

\bibitem{mitchell2022sampling}
Rory Mitchell, Joshua Cooper, Eibe Frank, and Geoffrey Holmes.
\newblock Sampling permutations for shapley value estimation.
\newblock 2022.

\bibitem{okhrati2021multilinear}
Ramin Okhrati and Aldo Lipani.
\newblock A multilinear sampling algorithm to estimate shapley values.
\newblock In {\em 2020 25th International Conference on Pattern Recognition
  (ICPR)}, pages 7992--7999. IEEE, 2021.

\bibitem{park2023trak}
Sung~Min Park, Kristian Georgiev, Andrew Ilyas, Guillaume Leclerc, and
  Aleksander Madry.
\newblock Trak: Attributing model behavior at scale.
\newblock 2023.

\bibitem{rombach2022high}
Robin Rombach, Andreas Blattmann, Dominik Lorenz, Patrick Esser, and Bj{\"o}rn
  Ommer.
\newblock High-resolution image synthesis with latent diffusion models.
\newblock In {\em Proceedings of the IEEE/CVF conference on computer vision and
  pattern recognition}, pages 10684--10695, 2022.

\bibitem{roth1988introduction}
Alvin~E Roth.
\newblock Introduction to the shapley value.
\newblock {\em The Shapley value}, pages 1--27, 1988.

\bibitem{sag2023copyright}
Matthew Sag.
\newblock Copyright safety for generative ai.
\newblock {\em Houston Law Review}, 61(2):295--347, 2023.

\bibitem{saleh2016large}
Babak Saleh and Ahmed Elgammal.
\newblock Large-scale classification of fine-art paintings: Learning the right
  metric on the right feature.
\newblock {\em International Journal for Digital Art History}, (2), 2016.

\bibitem{samuelson2023generative}
Pamela Samuelson.
\newblock Generative ai meets copyright.
\newblock {\em Science}, 381(6654):158--161, 2023.

\bibitem{shan2023glaze}
Shawn Shan, Jenna Cryan, Emily Wenger, Haitao Zheng, Rana Hanocka, and Ben~Y
  Zhao.
\newblock Glaze: Protecting artists from style mimicry by text-to-image models.
\newblock {\em arXiv preprint arXiv:2302.04222}, 2023.

\bibitem{shapley1953value}
Lloyd~S Shapley.
\newblock A value for $n$-person games.
\newblock {\em Contributions to the Theory of Games}, 2(28):307--317, 1953.

\bibitem{vyas2023provable}
Nikhil Vyas, Sham Kakade, and Boaz Barak.
\newblock Provable copyright protection for generative models.
\newblock {\em arXiv preprint arXiv:2302.10870}, 2023.

\bibitem{wang2023data}
Jiachen~T Wang and Ruoxi Jia.
\newblock Data banzhaf: A robust data valuation framework for machine learning.
\newblock In {\em International Conference on Artificial Intelligence and
  Statistics}, pages 6388--6421. PMLR, 2023.

\bibitem{wang2023notegroup}
Jiachen~T Wang and Ruoxi Jia.
\newblock A note on" towards efficient data valuation based on the shapley
  value''.
\newblock {\em arXiv preprint arXiv:2302.11431}, 2023.

\bibitem{wang2024data}
Jiachen~T Wang, Prateek Mittal, Dawn Song, and Ruoxi Jia.
\newblock Data shapley in one training run.
\newblock {\em arXiv preprint arXiv:2406.11011}, 2024.

\bibitem{wang2021improving}
Tianhao Wang, Yu~Yang, and Ruoxi Jia.
\newblock Improving cooperative game theory-based data valuation via data
  utility learning.
\newblock {\em ICLR 2022 Workshop on Socially Responsible Machine Learning},
  2022.

\bibitem{watt2014fair}
Richard Watt.
\newblock Fair remuneration for copyright holders and the shapley value.
\newblock {\em Handbook on the Economics of Copyright. A Guide for Students and
  Teachers}, 2014.

\bibitem{yang2023matching}
Jiaxi Yang, Wenglong Deng, Benlin Liu, Yangsibo Huang, and Xiaoxiao Li.
\newblock Matching-based data valuation for generative model.
\newblock {\em arXiv preprint arXiv:2304.10701}, 2023.

\bibitem{zheng2023intriguing}
Xiaosen Zheng, Tianyu Pang, Chao Du, Jing Jiang, and Min Lin.
\newblock Intriguing properties of data attribution on diffusion models.
\newblock {\em arXiv preprint arXiv:2311.00500}, 2023.

\end{thebibliography}
